\documentclass[runningheads,a4paper]{llncs}

\usepackage{amssymb}
\usepackage{amsmath}
\usepackage{graphicx}
\usepackage[caption=false]{subfig}
\usepackage{booktabs}
\usepackage{multirow}

\usepackage{url}
\urldef{\mailsa}\path|{alfred.hofmann, ursula.barth, ingrid.haas, frank.holzwarth,|
\urldef{\mailsb}\path|anna.kramer, leonie.kunz, christine.reiss, nicole.sator,|
\urldef{\mailsc}\path|erika.siebert-cole, peter.strasser, lncs}@springer.com|

\begin{document}

\mainmatter  % start of an individual contribution

%\title{Lecture Notes in Computer Science:\\Authors' Instructions
%for the Preparation\\of Camera-Ready
%Contributions\\to LNCS/LNAI/LNBI Proceedings}
\title{
%Intensity-based 2D/3D Registration via 
%Optimising CNN Pose Estimation on a Riemannian Manifold
Computing CNN Loss and Gradients for \\ Pose Estimation with Riemannian Geometry
}
\titlerunning{Computing CNN Loss and Gradients with Riemannian Geometry}

\author{Benjamin Hou$^{1}$, Nina Miolane$^{2}$, Bishesh Khanal$^{1,3}$, Matthew C.H. Lee$^{1,4}$, 
Amir Alansary$^{1}$, Steven McDonagh$^{1}$, Jo V. Hajnal$^{3}$, 
Daniel Rueckert$^{1}$, \\ Ben Glocker$^{1}$, and Bernhard Kainz$^{1}$}
%index{Hou, Benjamin}
%index{Alansary, Amir}
%index{McDonagh, Steven}
%index{Davidson, Alice}
%index{Rutherford, Mary}
%index{Hajnal, Jo V.}
%index{Rueckert, Daniel}
%index{Glocker, Ben}
%index{Kainz, Bernhard}
%
\authorrunning{B. Hou et al.}

\institute{
$^{1}$Imperial College London, $^{2}$INRIA \& Stanford, $^{3}$King’s College London, $^{4}$HeartFlow
%\url{https://biomedia.doc.ic.ac.uk}
}


\toctitle{Lecture Notes in Computer Science}
\tocauthor{Authors' Instructions}
\maketitle

\begin{abstract}

Pose estimation, \emph{i.e.} predicting a 3D rigid transformation with respect to a fixed co-ordinate frame in, $SE(3)$, is an omnipresent problem in medical image analysis. Deep learning methods often parameterise poses with a representation that separates rotation and translation. Available frameworks do not provide means to calculate loss on a manifold, regression is usually performed using the L2-norm independently on the rotation's and the translation's parameterisations. This is a metric for linear spaces that does not take into account the Lie group structure of $SE(3)$.  
We propose a general Riemannian formulation of the pose estimation problem, and train CNNs directly on $SE(3)$ equipped with a left-invariant Riemannian metric. The loss between the ground truth and predicted pose (elements of the manifold) is calculated as the Riemannian geodesic distance, which couples together the translation and rotation components. Network weights are updated by back-propagating the gradient with respect to the predicted pose on the tangent space of the manifold $SE(3)$. We thoroughly evaluate the effectiveness of our loss function by comparing its performance with popular and most commonly used existing methods, on tasks such as image-based localisation and intensity-based 2D/3D registration. We also show that hyper-parameters, used in our loss function to weight the contribution between rotations and translations, can be intrinsically calculated from the dataset to achieve greater performance margins. 

%\keywords{We would like to encourage you to list your keywords within the abstract section}
\end{abstract}

\section{Introduction}

Intensity-based registration and landmark matching are the de-facto standards to align data from multiple image sources into a common co-ordinate system. Applications that require intensity-based registration include \emph{e.g.}, atlas-based segmentation~\cite{Aljabar2009726}, motion-compensation~\cite{rousseau2006registration}, tracking~\cite{Miao2016}, or clinical analysis of the data visualised in a standard co-ordinate system. These often require manual initialisation of the alignment since general optimisation methods often cannot find a global minimum from any given starting position on the cost function. An initial rigid registration can be achieved by selecting common landmarks~\cite{Ghesu2016} through an iterative agent, which impedes hard real-time constraints or less robustly through local image descriptors \cite{ZITOVA2003977}. 

Convolutional Neural Networks (CNNs) have shown promising results for intra and inter modal alignment~\cite{Miao2016,Hou2017}. These approaches show that information about a learn-able canonical co-ordinate system is encoded directly in the features of an image. Early work in this domain showed that image's pose (i.e. position and orientation) can be regressed relatively to a canonical alignment from a large set of training images sampling the canonical space \cite{KendallGC15}. Follow-up formulations for medical applications showed similar success for motion compensation and device localisation~\cite{Miao2016,Hou2017}. However, these approaches rely on heuristic approximations and manual fine-tuning of the CNN loss used to characterise the poses' prediction error. This fosters domain shift problems and limits options for interchangeable application of various deep learning pose estimation models. 

\noindent\textbf{Contribution:} We introduce a new loss function that calculates the geodesic distance of two poses on the $SE(3)$ manifold, from a data-adaptive Riemannian metric. We derive appropriate gradients that are required for CNN back propagation. Our method couples the translation and rotation parameters, and regresses them simultaneously as one parameter on the Lie algebra $\mathfrak{se}(3)$. We show that our loss function is agnostic to the architecture by training different CNNs and can effectively predict poses that are comparable to state-of-the-art methods. In addition, we demonstrate that hyper-parameters tuning for our loss function can be directly calculated from the dataset, thus avoiding long and expensive optimisation searches to boost performance. Finally, we validate quantitatively by benchmarking the performance of our loss function with current state-of-the-art methods, and validate their statistical significance with Student's t-test. 

\noindent\textbf{Background:} A pose, i.e. a rigid transformation in 3D, is an element of the Lie group $SE(3)$, the Special Euclidean group in 3D. A pose has two components; a rotation component of group $SO(3)$ and a translation component of $\mathbb{R}^3$. $SE(3)$ has the following matrix representation (homogeneous representation):

\begin{equation}
SE(3) = \left\{ X \quad | \quad X = \left[ \begin{array}{c|c} R & t \\ \hline 0 & 1 \end{array} \right], t \in \mathbb{R}^3, R \in SO(3) \right\}
\end{equation}

In usual implementations of $SE(3)$, the rotation ($SO(3)$) can be parameterised in any form as long as the group structure is implicitly imposed. $R$ can be stored as Euler angles, quaternions, axis-angle or $SO(3)$ rotation matrix. The numerical properties of each parameterisation need to be considered carefully, especially when designing deep learning applications, as it can impact efficacy.

Hyunh et al.~\cite{huynh2009metrics} have shown that Euler parameterisation is not unique, this is undesirable as two different mappings can represent the same rotation. A rotation matrix, carrying 9 parameters, is over parameterised and has a strict ordinance on orthonormality. Non-orthogonal rotation matrix can result in skewed or sheared transformations, making it undesirable also. Quaternion parameterisation are often favoured as it can be mapped to valid transformations after normalisation. However, the parameterisation chosen for the rotation is almost never coupled with the parameterisation of the translation, thus denying the intrinsic structure of $SE(3)=SO(3) \ltimes \mathbb{R}^3$. Here, we choose to represent the rotation and the translation together as an element of the Lie algebra of $SE(3)$, i.e. its tangent space at the identity element of the group, denoted $\mathfrak{se}(3)$. It represents the best linear approximation of $SE(3)$ around its identity element. Since the Lie group $SE(3)$ is 6-dimensional, an element of $\mathfrak{se}(3)$ is a 6D vector.

One can define a collection of distances on $SE(3)$, which can be used as loss functions in deep learning applications. A popular choice for the loss is the Euclidean distance associated to the L2-norm. However, the L2-norm is not desirable on $SE(3)$ since it does not respect the manifold's non-linearity and can lead to unpredictable behaviours. It is also undesirable to use two separates L2-norms on $SO(3)$ and $\mathbb{R}^3$ since $SE(3)$ is not a direct product, and $SO(3)$ itself is non-linear: this can be observed visually with quaternions, e.g., the Euclidean distance of two quaternions can be small, despite the rotation being large. This disparity causes network weight updates to be sub optimal. Hence it is desirable to have a loss function that respects the structure of $SE(3)$.

\noindent\textbf{Related Work:} Popular deep learning frameworks, such as Caffe, TensorFlow, Theano, PyTorch, do not provide the means to regress on $SE(3)$, as the common losses used are cross-entropy for probabilities or a p-norms for distances.

Kendall et al.~\cite{KendallGC15} uses the L2-norm to regress parameters on the Lie algebra $\mathfrak{se}(3)$ directly, with a $\beta$ parameter to weight the contribution between rotation and translation. This was similarly performed by authors in~\cite{Pei2017,enlighten156798,Miao2016,liao2017artificial1}, who use the predicted parameters for registration tasks. Alternatively, \cite{kendall2017posenet,Diepol2012optimisation} re-parameterised the pose parameters as projected co-ordinates on a 2D view plane. This was similarly performed by~\cite{Hou2017,hou2018recon} with Anchor Points, where three arbitrary selected reference points on a 2D plane define the plane's location in 3D space. Using the L2-norm to calculate the Euclidean distance between a predicted projection co-ordinate and the ground truth projection co-ordinate is appropriate, as the L2-norm is the appropriate metric. To the best of our knowledge, there is currently no loss function that respects the full Lie group structure of $SE(3)$, for example invariant Riemannian metrics on $SE(3)$ have not been used.

\section{Method}

\begin{figure}[htb]
\centering
\includegraphics[width=0.6\linewidth]{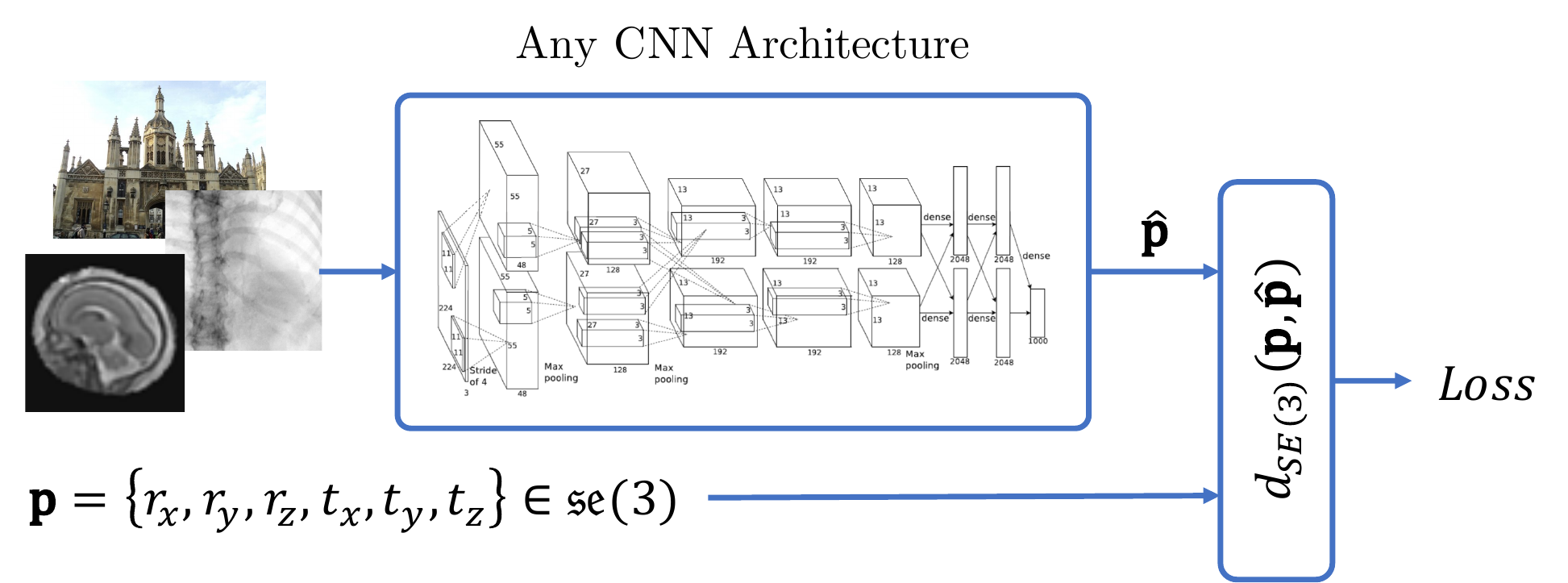}
\caption{CNN architecture using a Riemannian geodesic distance on $SE(3)$ as the loss.}
\label{fig:overview}
\end{figure}

The core of our method is to implement a new loss layer that is agnostic to the network architecture used: we define the loss as the geodesic distance on $SE(3)$ equipped with a left-invariant Riemannian metric, shown in Figure~\ref{fig:overview}.

\noindent\textbf{Left-invariant Riemannian metric on $SE(3)$:} A Riemannian metric on $SE(3)$ is a smooth collection of positive definite inner products on each tangent space of $SE(3)$. Then, $SE(3)$ becomes a Riemannian manifold. With a left-invariant metric, it is enough to define an inner product on the tangent space at the identity of $SE(3)$, and then ``propagate" it: the metric is s.t. $\forall u, v \in T_{p_1}SE(3)$ and $\forall p_1,p_2 \in SE(3)$:
$
<DL_{p_1}(p_2) u, DL_{p_1}(p_2) v>|_{L_{p_1} p_2} = <u, v>|_{p_2}
$
where $L_{p_1}$ is the left translation by $p_1$: $L_{p_1}(p_2) = p_1 \circ p_2$, and $DL_{p_1}(p_2)$ its differential at $p_2$. We define an inner product $Z$ at $p_2 = \text{identity}$ so that we get a metric $Z_{p_1}$ at the tangent space of any pose $p_1$ of $SE(3)$~\cite{miolane:hal-01133922}, and thus to compute inner products and norms of tangent vectors at $p_1$.

\noindent\textbf{Loss and gradient:} We use the loss function:
$
\text{loss}(\mathbf{p},\hat{\mathbf{p}})
    = \text{dist}_{SE(3)}^Z(\mathbf{p},\hat{\mathbf{p}})^2
    = {\left\| \text{Log}_{\hat{\mathbf{p}}}^{Z}(\mathbf{p}) \right\| }_{Z_{\hat{\mathbf{p}}}}^2 
$
where $\text{dist}_{SE(3)}^Z$ is the geodesic distance and Log is the Riemannian logarithm at $\hat{\mathbf{p}}$ i.e. a tangent vector at $\hat{\mathbf{p}}$. We use a left-invariant Riemannian metric, thus: $\text{loss}(\mathbf{p},\hat{\mathbf{p}}) =  ||DL_{\hat{\mathbf{p}}^{-1}}.\text{Log}_{\hat{\mathbf{p}}}^{Z}(\mathbf{p})||_{Z}^2$, where we now have a tangent vector at the identity and we can use the inner product $Z$. If we take $Z$ being the canonical inner product at identity, this is the L2-norm but on the tangent vector transported from $\hat{\mathbf{p}}$ to identity using the differential $DL_{\hat{\mathbf{p}}^{-1}}$. The backward gradient corresponding to the loss seen as a function of $\hat{\mathbf{p}}$ is
$
%\begin{equation}
    \nabla_{\hat{\mathbf{p}}} \text{loss}(\mathbf{p},\hat{\mathbf{p}}) = -2 \cdot \text{Log}_{\hat{\mathbf{p}}}^{Z}({\mathbf{p}})
%\end{equation}
$ \cite{pennec1999probabilities} which is a tangent vector at $\hat{\mathbf{p}}$.

\noindent\textbf{Implementation:} The inputs to the loss layer are the poses $\mathbf{p}$ and $\hat{\mathbf{p}}$ for ground truth and prediction respectively. We represent a pose with \texttt{geomstats} implementation \cite{miolane_2018} \emph{i.e.} as the Riemannian Logarithm of canonical left-invariant metric on $SE(3)$ s.t. $ p = \{ r, t \}  = \{ r_x , r_y , r_z , t_x , t_y , t_z \} \in \mathbb{R}^6$. With this parameterisation, the rotation $r$ is in axis-angle parameterisation, the inner product $Z$ is a 6x6 positive definite matrix and the differential $DL_{\hat{p}}$ of the left translation is the 6x6 jacobian matrix: 
$
J_{\hat{p}} = \begin{pmatrix}
\frac{\partial {L_{\hat{p}}}^{r}}{\partial r}
        & \frac{\partial {L_{\hat{p}}}^{r}}{\partial t} \\
\frac{\partial {L_{\hat{p}}}^{t}}{\partial r}
        & \frac{\partial {L_{\hat{p}}}^{t}}{\partial t}
\end{pmatrix}
$. We denote $v_t = \text{Log}_{\hat{\mathbf{p}}}^{Z}(\mathbf{p})$ which is a tangent vector at $\hat{\mathbf{p}}$ in this parameterisation. The loss is calculated by $\text{loss}(\mathbf{p},\hat{\mathbf{p}}) = v_t^T \ast J_{\hat{p}^{-1}}^T \ast Z \ast J_{\hat{p}^{-1}} \ast v_t$ where $*$ is the matrix multiplication and the Riemannian logarithm $v_t$ is given by \texttt{geomstats}. The gradient is calculated by: $\nabla_{\hat{\mathbf{p}}} \text{loss}(\mathbf{p},\hat{\mathbf{p}})= -2 * J_{{\hat{p}}^{-1}}^T \ast Z \ast J_{{\hat{p}}^{-1}} *  v_t$.

\section{Experiments and Results}

We evaluate our novel loss function on three datasets: \textbf{(Exp1)} the publicly available PoseNet dataset~\cite{KendallGC15}, which allows a direct comparison to state-of-the-art in Computer Vision and further evaluates optimisation strategies for these experiments. \textbf{(Exp2)}, C-Arm X-Ray to Computed Tomography (CT) alignment problem with data from~\cite{Hou2017}. \textbf{(Exp3)}, the pose estimation dataset for motion compensation in fetal Magnetic Resonance Imaging (MRI) from~\cite{hou2018recon}. 

In each experiment, we benchmark existing $SE(3)$ parameterisation strategies with the respective loss function used. PoseNet: direct regression of parameters on the Lie algebra $\mathfrak{se}(3)$, where a combination of quaternion and translation parameters are regressed using L2-norms with a static $\beta$ parameter to weight the respective contribution. Anchor Points formulation: a re-parameterisation of $SE(3)$ in Euclidean space, where three statically defined points in 3D space defines a plane. Each Anchor Point is regressed independently using the L2-norm. Finally, our $SE(3)$ loss, i.e., the geodesic distance on the Riemannian manifold. 
All experiments are conducted using the Caffe framework, on a computer equipped with an Intel i7 6700K CPU and Nvidia Titan X Pascal GPU.

\noindent\textbf{Exp.1: Metric Localisation on Natural Images:} In this experiment, we replicated PoseNet's original experiment~\cite{KendallGC15} on the King's College dataset as a baseline benchmark. \cite{KendallGC15} extracted images from a series of videos, and fed them into a structure from motion pipeline to create a 3D model in order to extract plane locations with respect to a world co-ordinate reference frame. The parameterisations of this dataset are quaternions with translation offsets. We mirrored the dataset using axis-angle representation instead of quaternions, and used our $SE(3)$ loss function as regressor. Both networks were trained with a GoogLeNet~\cite{googlenet} base architecture with no parameter weighting. %Table~\ref{tab:PoseNet-SE3-no-weights} summarises the results. 

\begin{table}[htb]
\centering
\caption{Mean Error of Loss Functions on Natural Images}
\label{tab:PoseNet-SE3}
\begin{tabular}{c|cccccc|c}
\toprule
~~~~~~~~~~~~~~~~ & ~~~$R_x$~~~ & ~~~$R_y$~~~ & ~~~$R_z$~~~ & ~~~$t_x$~~~ & ~~~$t_y$~~~ & ~~~$t_z$~~~ &   ~~~G.D.~~~  \\
\midrule
PoseNet          & \textbf{4.141} & 7.774           & \textbf{4.597} & 1.341            & \textbf{1.139}    & \textbf{0.154}    &  23.629           \\
$SE(3)$           & 4.306          & \textbf{6.675}  & 11.580         & \textbf{1.307}   & 1.149             & 0.155             &  \textbf{14.973}  \\       
\bottomrule
\multicolumn{8}{c}{A: Without Parameter Weighting} \\
\toprule
~~~~~~~~~~~~~~~~ & ~~~$R_x$~~~ & ~~~$R_y$~~~ & ~~~$R_z$~~~ & ~~~$t_x$~~~ & ~~~$t_y$~~~ & ~~~$t_z$~~~ &   ~~~G.D.~~~  \\
\midrule
PoseNet          & \textbf{1.790}   & \textbf{2.612} & \textbf{2.371}   & \textbf{1.161} & 1.306          & \textbf{0.154}  &  \textbf{13.516}  \\
$SE(3)$          & 1.870            & 3.143          & 3.662            & 1.759          & \textbf{1.240} & 0.156           &  16.370           \\       
\bottomrule
\multicolumn{8}{c}{B: With Parameter Weighting}
\end{tabular}
\end{table}

We convert the predicted and ground truth poses to Euler angles in degrees and translation in meters, along with the geodesic distance (G.D.) on the manifold. Table~\ref{tab:PoseNet-SE3}-A shows the average errors of each parameter. It can be seen that the error is similar in each Euler and translation parameters, which is confirmed by Student's t-test to be insignificant. However, the geodesic distance error of SE(3) is much lower compared to PoseNet. Despite this, Student's t-test still shows no significant difference, which is caused by the large variance of PoseNet. 

To tune the weight parameter $\beta$, Kendall et al.~\cite{KendallGC15} performed grid search and found that $\beta = [120,750]$ works best for indoor scenes and $\beta = [250,2000]$ for outdoor scenes. Grid search is computationally very expensive, and it can be difficult to find an optimal value if the search interval is coarse. We show here that we can compute a data-adaptive Riemannian metric on $SE(3)$ to weight the contribution of each parameter in the loss. 

We first train the network with no weightings, followed by an inference pass through the entire validation dataset. We compute the prediction error as the rigid transformation: ${(y_{true})}^{-1} \circ y_{pred}$ and consider the dataset of their Riemannian logarithms at the identity $\{X_i\}_i$. The parameter weightings are then calculated by $\text{diag}({\text{cov}(X_i)}^{-1})$. The diagonal of the covariance matrix shows the variance of each parameter, whereas the inverse shows how tightly coupled it is to the mean. Thus, the higher the diagonal element, the tighter the variable is clustered. Elements, that are more sparsely coupled, are weighted less as they are likely to induce errors. The optimal weightings for the King's College dataset from \cite{KendallGC15} are: 
$
%\begin{equation}
    \text{diag}({\text{cov}(X_i)}^{-1}) = \{ 0.147 , 0.954 , 0.261 , 0.001 , 0.003 , 0.002  \}
%\end{equation}
$. 

% 0.14785401,  0.95406175,  0.26056108,  0.00101328,  0.00311928,  0.00159111

Table~\ref{tab:PoseNet-SE3}-B shows the performance of the networks retrained with suggested weightings. 
%Benchmarking with Geodesic Distances is not performed here, as it is not possible to map hyper-parameters across different metrics. The newly trained networks shows significantly improved performance compared to the networks trained without parameter weighting. However 
Student's T-test still shows no significant difference between errors induced by PoseNet and $SE(3)$. We note that having different weightings on the rotation part induces a distance that is not a Riemannian geodesic distance anymore. The properties of this distance will be investigated in future work.

\noindent\textbf{Exp.2: Plane Detection on C-ARM Imaging:} \cite{Hou2017} demonstrated the versatility of CNNs for performing 2D/3D registration of C-Arm X-ray images to CT volumes using Anchor Points. In this experiment, we replicate the 2D/3D registration task using CaffeNet and evaluate the performance with the newly proposed $SE(3)$ parameterisation and loss regressor. For comparison, a network was also trained with PoseNet's parameterisation. All weight parameters are set to a default of 1. Table~\ref{tab:DRR-results} shows the performance of each parameterisation. 

\begin{table}[htb]
\centering
\caption{Mean Error of Loss Functions on DRR (Digitally Reconstructed Radiographs)}
\label{tab:DRR-results}
\begin{tabular}{c|cccccc|c}
\toprule
~~~~~~~~~~~~~~~~~~~~~~ & ~~~~$R_x$~~~~ & ~~~~$R_y$~~~~ & ~~~~$R_z$~~~~ & ~~~~$t_x$~~~~ & ~~~~$t_y$~~~~ & ~~~~$t_z$~~~~ &  ~~~~G.D.~~~~   \\
\midrule
PoseNet                & 7.960          & 3.136          & 7.547          & 62.650          & 57.315          & 45.852          &  15201.845         \\
Anchor Points          & \textbf{7.274} & \textbf{2.511} & \textbf{7.059} & 59.292          & \textbf{54.889} & \textbf{40.576} &  15115.858         \\
$SE(3)$                & 8.243          & 3.697          & 7.924          & \textbf{58.647} & 55.477          & 44.189          & \textbf{14170.722}* \\       
\bottomrule
\multicolumn{8}{c}{A: Healthy Patient Dataset}                                                                                \\
%\multicolumn{8}{c}{}                                                                                                       \\ 
\toprule
~~~~~~~~~~~~~~~~~~~~~~ & ~~~$R_x$~~~ & ~~~$R_y$~~~ & ~~~$R_z$~~~ & ~~~$t_x$~~~ & ~~~$t_y$~~~ & ~~~$t_z$~~~ &  ~~~G.D.~~~   \\
\midrule
PoseNet                & 10.653         & 5.788          & 10.760         & 69.107          & 72.238          & 57.726          &  23495.708           \\
Anchor Points          & \textbf{8.540} & \textbf{4.060} & \textbf{8.553} & 65.521          & \textbf{68.543} & 54.133          &  21725.921           \\
$SE(3)$                & 10.511         & 6.789          & 11.913         & \textbf{62.588} & 68.747          & \textbf{54.110} &  \textbf{19624.246}* \\       
\bottomrule
\multicolumn{8}{c}{B: Pathological Patient Dataset}                                                          
\end{tabular}
\end{table}

Here, we convert the predicted and ground truth poses to Euler angles in degrees, and translation to millimeters. Similar to Exp1, the average error for rotation and translation (for both healthy and pathological patients) are similar, and insignificant as confirmed by Student's t-test. However, there is a noticeable trend in average geodesic distance errors. Student's t-test showed significant difference (marked by *) between $SE(3)$ loss compared to PoseNet and Anchor Points for both datasets. This shows that the geodesic metric is able to quantify properties that the metric expressed in Euler-translation parameters cannot. 

\noindent\textbf{Exp.3: 2D/3D Registration on fetal brain MRI:} We replicate the results evaluation method from~\cite{hou2018recon}, and evaluate our loss regressor for 2D/3D registration used during motion compensation of fetal MRI data in canonical organ space. \cite{hou2018recon} uses aligned, reconstructed 3D brain volumes to learn a canonical orientation space and utilises an approach based on GoogLeNet to reorient unseen 2D brain slices into their correct anatomical location in this space. To sample the canonical training space we use the same Euler iteration method (\emph{i.e.}, $18^\circ$ steps in $R_x$ $R_y$ $R_z$ between $-90^{\circ}$ and $+90^{\circ}$, and 2mm offsets in $T_z$ constrained between -40 and 40) to generate 1.12M images for the training set. The evaluation method is performed similarly as~\cite{hou2018recon}, with the performance summarised in Table~\ref{tab:exp3}. The validation dataset is composed  of brain slices sampled with random Euler angles between $-90^\circ$ and $+90^\circ$, and random offsets between -40 and +40. 

\begin{table}[ht]
\centering
\caption{Mean Error of Loss Functions on Fetal Brain Images}
\label{tab:exp3}
\begin{tabular}{c|cccc|c}
\toprule
~~~~~~~~~~~~~~~~~~~~~~ & ~~~~~CC~~~~~    & ~~~~MSE~~~~       & ~~~PSNR~~~       & ~~~SSIM~~~      & ~~~~~G.D.~~~~~     \\ 
\midrule
%Eul-Cart               & 0.8148          & 1077.8            & 18.5254          & 0.537           & 18.749             \\
PoseNet                & 0.8199          & 1046.4            & 18.6509          & 0.5448          & 18.1708            \\ 
Anchor Points          & 0.8378          & 935.0             & 19.3564          & 0.5845          & 15.7504            \\
$SE(3)$                & \textbf{0.8732}* & \textbf{724.9713}* & \textbf{20.7484}* & \textbf{0.6470}* & \textbf{10.0836}*   \\
\bottomrule
\end{tabular}
\end{table}

Our $SE(3)$ loss function shows drastic improvement in all image similarity metrics (Cross Correlation, Mean Squared Error, Peak Signal-to-Noise Ratio and Structural SIMilarity). This is confirmed by Student's t-test which shows significant difference. This is crucial for Slice-to-Volume Registration (SVR) applications as the metric for slice alignment is derived from the metrics above~\cite{kainz2015fast}. 

\textbf{Discussion:} A pose is a combination of rotation and translation, therefore it seems reasonable that a CNN predicting a pose should use a metric that accounts for both simultaneously. Metrics are perceptually a method of measurement with its own set of rules, \emph{e.g.}, imperial vs. metric system for quantifying distances. Choosing a metric for a target application is not always straight forward and often a question of required precision, \emph{e.g.}, one would not measure the diameter of a pinhead with a meter rule, nor measure distance between cities with a caliper. We have shown that our loss function, using a Riemannian geodesic distance on $SE(3)$ is better suited for medical registration tasks as shown in Exp2 and Exp3. Exp2 shows each test case yielding no significant difference on Euler and translation parameters, with significant difference on geodesic parameters. This suggests that Euler-translation parameters separately are not able to fully quantify the properties of $SE(3)$. In Exp3, our loss function was able to significantly improve the image similarity metrics, as used by SVR algorithms.

\section{Conclusion}

In this work, we have presented a novel loss function to regress poses on the Lie group $SE(3)$, and derived the necessary gradients required for CNN training. We showed that our method alleviates the need of re-parameterising regression parameters, which addresses the domain shift problem of deep learning applications. Our approach achieves similar results to manually fine-tuned approximations out-of-the-box,  \emph{e.g.}, for data from a new scanner. This is demonstrated on the current state-of-the-art for pose estimation, PoseNet, where we show that our method achieves similar performance as the carefully tuned approximation used in \cite{KendallGC15}. We also show significant improvements for medical image pose estimation and outperform the state-of-the-art in this domain~\cite{Hou2017,hou2018recon}. \textbf{Acknowledgements:} Supported by the Wellcome Trust IEH Award [102431] and Nvidia.

\bibliographystyle{splncs03}
\bibliography{references}

\end{document}